\documentclass[10pt,twocolumn,a4paper,fleqn]{article}
\usepackage[a4paper,left=15mm,right=15mm,top=30mm,bottom=25mm,columnsep=10mm]{geometry}
\usepackage{graphicx}
\usepackage{times}
\usepackage{titlesec}
\usepackage{indentfirst}
\usepackage[fleqn]{amsmath}
\usepackage{fancyhdr}
\usepackage{cite}
\usepackage{enumitem}
\usepackage{etoolbox}
\renewcommand{\normalsize}{\fontsize{10pt}{11.9pt}\selectfont} 
\titleformat{\section}
  {\normalfont\large\bfseries\MakeUppercase}{\thesection.}{0.5em}{}
\titlespacing*{\section}{0pt}{2.5ex plus 0.5ex minus .2ex}{1.5ex}
\titleformat{\subsection}
  {\normalfont\normalsize\bfseries}{\thesubsection.}{0.5em}{}
\titlespacing*{\subsection}{0pt}{1.5ex}{0.5ex}

\titleformat{\subsubsection}
  {\normalfont\normalsize}{\thesubsubsection.}{0.5em}{}
\titlespacing*{\subsubsection}{0pt}{2.0ex plus .5ex minus .2ex}{0.1ex}

\setlist[description]{
  font=\itshape,
  nosep,
  leftmargin=0pt,
  labelsep=0.5em
}

\AtBeginEnvironment{table}{
  \scriptsize
}

\usepackage[small,bf]{caption}
\DeclareCaptionLabelSeparator{periodspace}{.~ }
\makeatletter
\newcommand{\affils}[1]{\def\@affils{#1}}
\renewcommand{\abstract}[1]{\def\@abstract{#1}}
\newcommand{\keywords}[1]{\def\@keywords{#1}}

\renewcommand{\@maketitle}{%
  \newpage
  \null
  \begin{center}
    {\fontsize{15pt}{15pt}\selectfont \bfseries \@title \par}
    \vskip 1.0em
    {\large \@author \par}
    \vskip 0.5em
    {\normalsize \@affils \par}
  \end{center}
  {\normalsize \noindent \textbf{Abstract:} \@abstract \par}
  \vskip 1em
  {\normalsize \noindent \textbf{Keywords:} \@keywords \par}
  \vskip 1.4em
}
\makeatother

\makeatletter
\renewenvironment{thebibliography}[1]{
  \section*{\refname}
  \normalsize
  \list{[\arabic{enumi}]}{
    \settowidth\labelwidth{[#1]}
    \leftmargin\labelwidth
    \advance\leftmargin\labelsep
    \setlength{\itemsep}{0pt}
    \setlength{\parsep}{0pt}
    \setlength{\topsep}{0pt}
    \setlength{\partopsep}{0pt}
    \usecounter{enumi}
    
  }
  
  \sloppy\clubpenalty4000\widowpenalty4000
  \sfcode`\.=1000\relax
}{
  \endlist
}
\makeatother

\usepackage{subcaption}
\usepackage{amsmath}
\usepackage{multirow}
\usepackage{url}
\allowdisplaybreaks

\newcommand{\mypara}[1]{\vspace{1.0mm}\noindent\textbf{#1}.\hspace{1.0mm}}

\fancypagestyle{firstpage}{
  \fancyhf{}
  
  \fancyfoot[C]{\footnotesize \sffamily Author's version. In Proc. of the Joint Symposium of AROB 31st and ISBC 11th (AROB-ISBC 2026), pp.~787--792, 2026.}
}

\title{Adaptive Policy Switching of Two-Wheeled Differential Robots\\for Traversing over Diverse Terrains}

\author{Haruki Izawa${}^{1}$, Takeshi Takai${}^{2}$, Shingo Kitano${}^{2}$, Mikita Miyaguchi${}^{2}$, and Hiroaki Kawashima${}^{1}$}

\affils{${}^{1}$Graduate School of Information Science, University of Hyogo, Hyogo, Japan\\
(E-mail: izawa.hyogo@gmail.com, kawashima@gsis.u-hyogo.ac.jp)\\
${}^{2}$Takenaka Research \& Development Institute, Takenaka Corporation, Chiba, Japan\\
(E-mail: \{takai.takeshi, kitano.shingo, miyaguchi.mikita\}@takenaka.co.jp)\\
}
\abstract{
Exploring lunar lava tubes requires robots to traverse without human intervention. Because pre-trained policies cannot fully cover all possible terrain conditions, our goal is to enable adaptive policy switching, where the robot selects an appropriate terrain-specialized model based on its current terrain features. This study investigates whether terrain types can be estimated effectively using posture-related observations collected during navigation. We fine-tuned a pre-trained policy using Proximal Policy Optimization (PPO), and then collected the robot's 3D orientation data as it moved across flat and rough terrain in a simulated lava-tube environment. Our analysis revealed that the standard deviation of the robot's pitch data shows a clear difference between these two terrain types. Using Gaussian mixture models (GMM), we evaluated terrain classification across various window sizes. An accuracy of more than 98\% was achieved when using a 70-step window. The result suggests that short-term orientation data are sufficient for reliable terrain estimation, providing a foundation for adaptive policy switching.
}


\begin{document}

\maketitle
\thispagestyle{firstpage}

\section{Introduction}
The space industry evolves so rapidly that we have actual plans to develop other planets such as the Moon~\cite{moon_exp} and Mars to get natural resources or pursue the truth of the universe. In this situation, exploring and developing these unexplored regions requires the use of robots much more to mitigate risks and reduce costs. In fact, rovers such as Curiosity~\cite{mars_cur} and Perseverance~\cite{mars_per} are used in the Martian exploration. However, direct human operation from Earth is challenging due to the large physical distance. Furthermore, when the environment of such unexplored areas is underground such as lava tubes~\cite{lavatube}, making direct observation by probes impossible, it is difficult to accurately grasp the terrain and other conditions beforehand. It is also challenging to predetermine how the robot should act on-site in this situation. For these reasons, robots tasked with exploring and developing unexplored areas need to make decisions and act based on their observations without human intervention.

Reinforcement learning, which does not require human intervention, is one of the effective ways in this field because it enables self-decision-making. However, such decision-making is possible only in environments seen during training, and it becomes difficult to achieve in unexpected environments. The unexplored areas targeted in this research are assumed to contain a mixture of various unknown terrains, making it difficult to anticipate all terrain conditions in advance or to acquire policies capable of handling such diversity beforehand. Therefore, our long-term goal is to build an approach that pools policy models for various terrains as resources in advance, enabling model switching among them during operation.

In this paper, we propose a method to capture terrain features using the observations the robot is expected to obtain. The experiments are conducted in the two kinds of terrain; one is the flat area and the other is the rough area. First, we trained a policy model in both the flat and rough areas. The robot then traversed these two types of terrain using the pre-trained general model and collected its 3D orientation data. We analyzed these data and evaluated how effectively the robot can estimate the terrain features of the area.

\section{Related work}

\subsection{Proximal Policy Optimization}
Proximal Policy Optimization (PPO)~\cite{Schulman_PPO_2017} is a reinforcement learning method that achieves stable learning by limiting how much the policy can change at each update through clipping. One advantage of policy-based algorithms like PPO is that they can handle continuous action spaces.
In our study, we also use PPO instead of other reinforcement learning algorithms such as DQN (Deep Q-network)~\cite{Mnih_dqn_2013}, REINFORCE~\cite{williams_reinforce_1992}, or DDQN (Double DQN)~\cite{van_ddqn_2016} , because it can work with continuous actions and provides stable learning performance, which is important for controlling the two-wheeled differential robot.

\subsection{Cooperative exploration on the Moon}
In future lunar missions, several robots are expected to explore together. Some studies use CNNs to extract useful information from observations and decide where each robot should move.
Yu et al. proposed Asynchronous Coordination Explorer (ACE) ~\cite{Yu_AAMAS2023}, which makes multi-robot exploration more efficient. Standard multi-agent PPO (MAPPO) requires all robots to complete their actions before moving to the next step, which slows down exploration. To solve this, Yu et al. used Async-MAPPO, where each robot can act without waiting for others. They also added Action-Delay Randomization to handle communication delays and used a Multi-Tower-CNN-Based Policy to speed up processing.

\section{Adaptive policy switching}
\subsection{Overview}
The terrain-specialized models have the potential to enable smoother navigation across different areas compared with models that cover mixed terrain types~\cite{Izawa_IPJS2025}.  
Table~\ref{tab:result_ipsj} shows the target-reaching success rates and average target-reaching times for different combinations of models and terrains, suggesting the effectiveness of terrain-specialized models. 
For traversing over diverse terrains, we therefore address a problem of adaptive policy switching. 
In particular, we envision a system in which robots maintain a pool of models specialized for different terrains as resources and select the most effective one for the terrain they encounter.

To train a model specialized for a particular terrain type, the robot needs to learn within an environment that exhibits similar terrain conditions. However, since these models are trained autonomously on-site, the robot must identify terrain features without human supervision. For example, when the robot is in a flat area, it should train or update a model specialized for flat terrain only in that flat region. Without identifying the terrain features, the robot may train it both in flat and rough areas, resulting in a model that is no longer specialized for the flat terrain. The ability to identify terrain features without human intervention enables the robot to determine the terrains it is currently traversing, where it should conduct model training, and what type of terrain the model is specialized for.

\begin{figure}[tbp]
    \centering
    \includegraphics[width=0.9\linewidth]{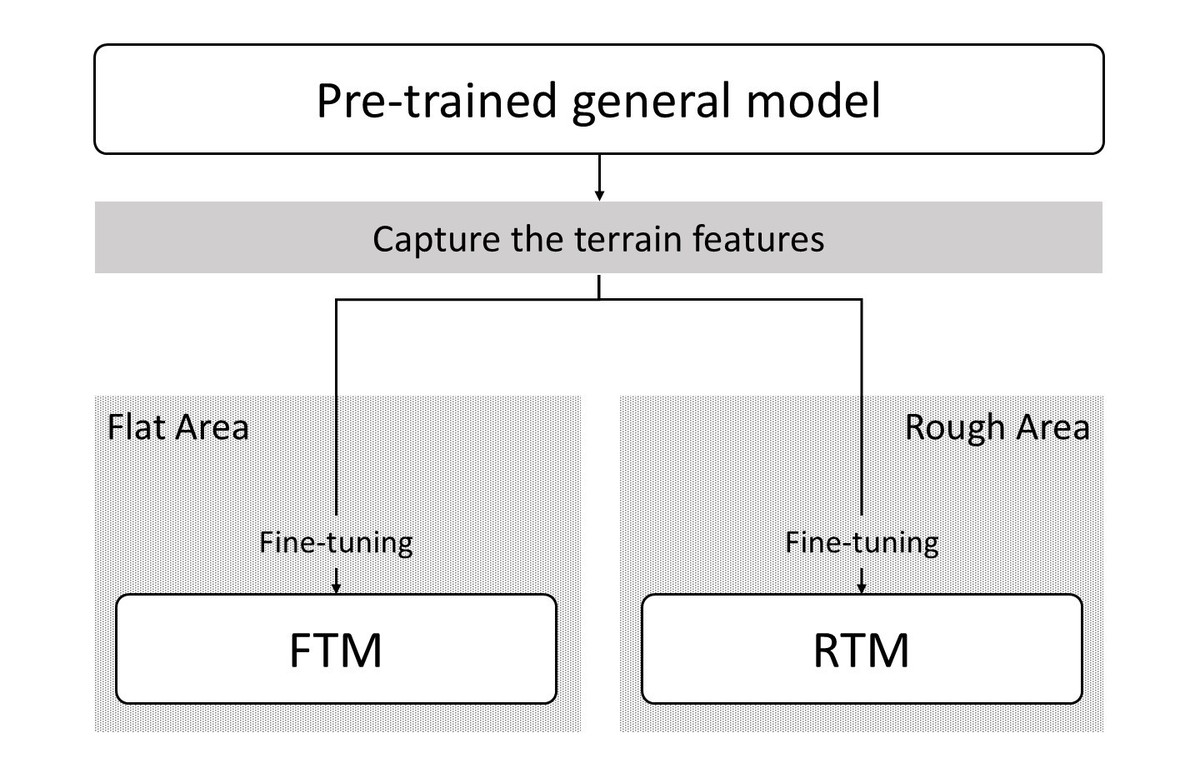}
    \caption{Training process for terrain-specialized models}
    \label{fig:finetuning}
    % \vspace{6mm}
\end{figure}

Figure~\ref{fig:finetuning} illustrates the training process for terrain-specialized models, such as the flat terrain model (FTM) and the rough terrain model (RTM), obtained after terrain identification. We first train a general model, which is used for terrain identification and subsequently serves as the initialization for fine-tuning toward terrain-specialized models.
In this paper, we focus on the terrain identification stage, where we obtain a pre-trained general model to extract terrain features useful for terrain identification prior to fine-tuning.

\begin{table}[bp]
    \caption{Target-reaching success rates and average target-reaching times with the combination of models and terrains~\cite{Izawa_IPJS2025}. A general model is trained in both flat and rough areas.}
    \label{tab:result_ipsj}
    \begin{subfigure}[t]{\linewidth}
    \centering
    \caption{Target-reaching success rate (\%): Numbers in parentheses indicate the number of arrivals out of 50 trials}
    \begin{tabular}{l|rr}  %
    \hline
    Policy model & Flat area & Rough area \\
    \hline  %
    Flat terrain & 98.0 (49) & 76.0 (38) \\
    Rough terrain & 98.0 (49) & \textbf{96.0 (48)}  \\
    General & \textbf{100.0 (50)} & 88.0 (44)  \\
    \hline
    \end{tabular}
    \end{subfigure}

    \vspace{5mm}

    \begin{subfigure}[t]{\linewidth}
    \caption{Average target-reaching time (s): Standard deviation in parentheses~\cite{Izawa_IPJS2025}}
    \centering
    \begin{tabular}{l|rr}  %
    \hline
    Policy model & Flat area & Rough area \\
    \hline  %
    Flat terrain & \textbf{5.47 (1.64)} & 9.63 (4.29) \\
    Rough terrain & 8.33 (5.92) & \textbf{8.96 (5.74)} \\
    General  & 5.90 (2.54) & 10.51 (7.35) \\
    \hline
    \end{tabular}
    \end{subfigure}
\end{table}

\subsection{Environment}
\label{sec:environment}

\begin{figure}[tbp]
    \centering

        \includegraphics[width=0.9\linewidth]{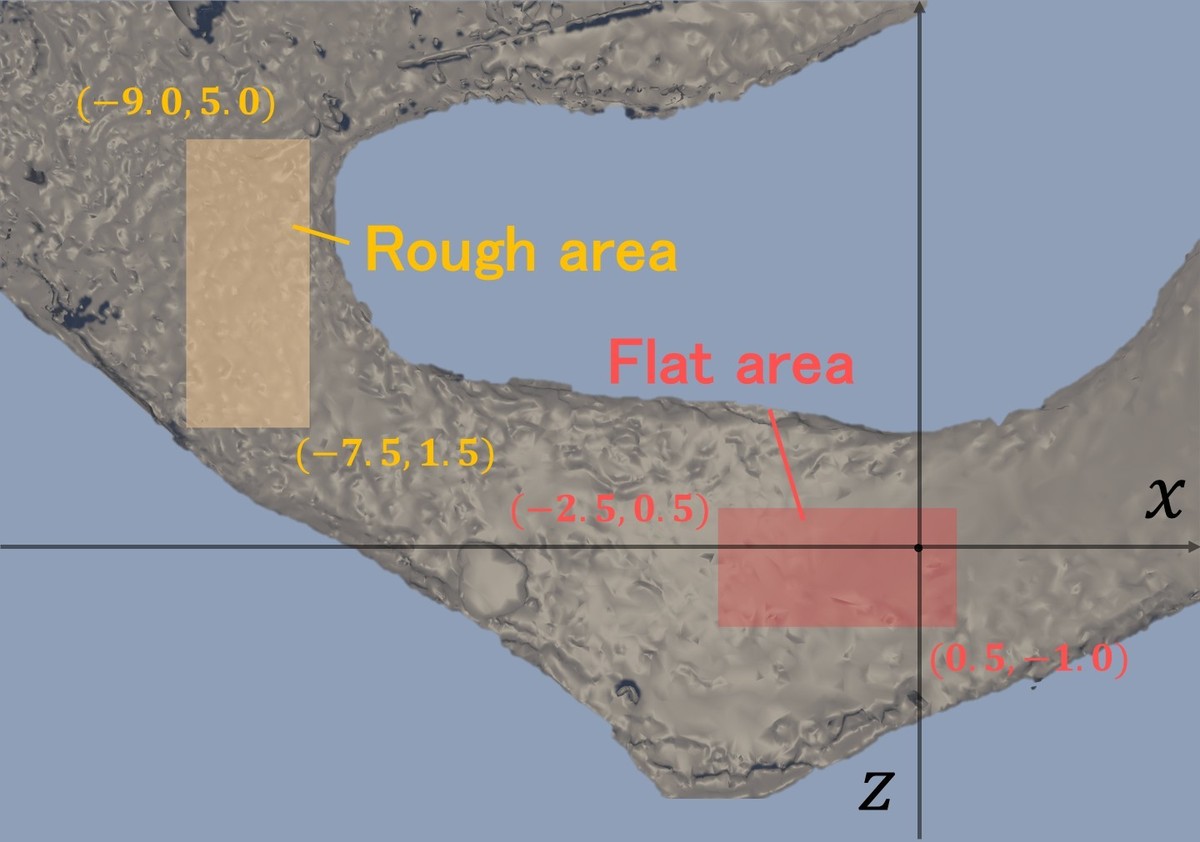}
        \caption{Simulation environment}
        \label{fig:mstube}

\end{figure}

Recently, potential lava-tube skylights have been discovered on the lunar surface. 
These lava tubes are expected to provide protection from X-rays and other harmful environmental factors, making them suitable locations for constructing bases.
Because establishing bases within such lava tubes using swarm robots is a key long-term objective in our project, this study focuses on simulations with terrain data obtained from a lava tube on the Earth.
To conduct simulations, we developed a Unity environment modeled after the Lake Sai Bat Cave located in Fujikawaguchiko Town, Yamanashi, Japan. We selected flat and rough areas from this environment. As shown in Fig.~\ref{fig:mstube}, we defined the origin of the world coordinate system around the center of the flat area, with the X-axis and Z-axis as the horizontal plane and the Y-axis as the vertical upward direction.
We defined the flat area as the region enclosed clockwise from the upper right by $(x, z) = (0.5, 0.5), (0.5,-1.0), (-2.5,-1.0), (-2.5, 0.5)$, and the rough area as the region enclosed by $(x, z) = (-7.5, 5.0), (-7.5, 1.5), (-9.0, 1.5), (-9.0, 5.0)$ (unit: meter). Note that the rough area's surface unevenness was reduced to 80\% of its original level.

\subsection{Robot}
\begin{figure}[tbp]
    \centering
    \includegraphics[width=0.8\linewidth]{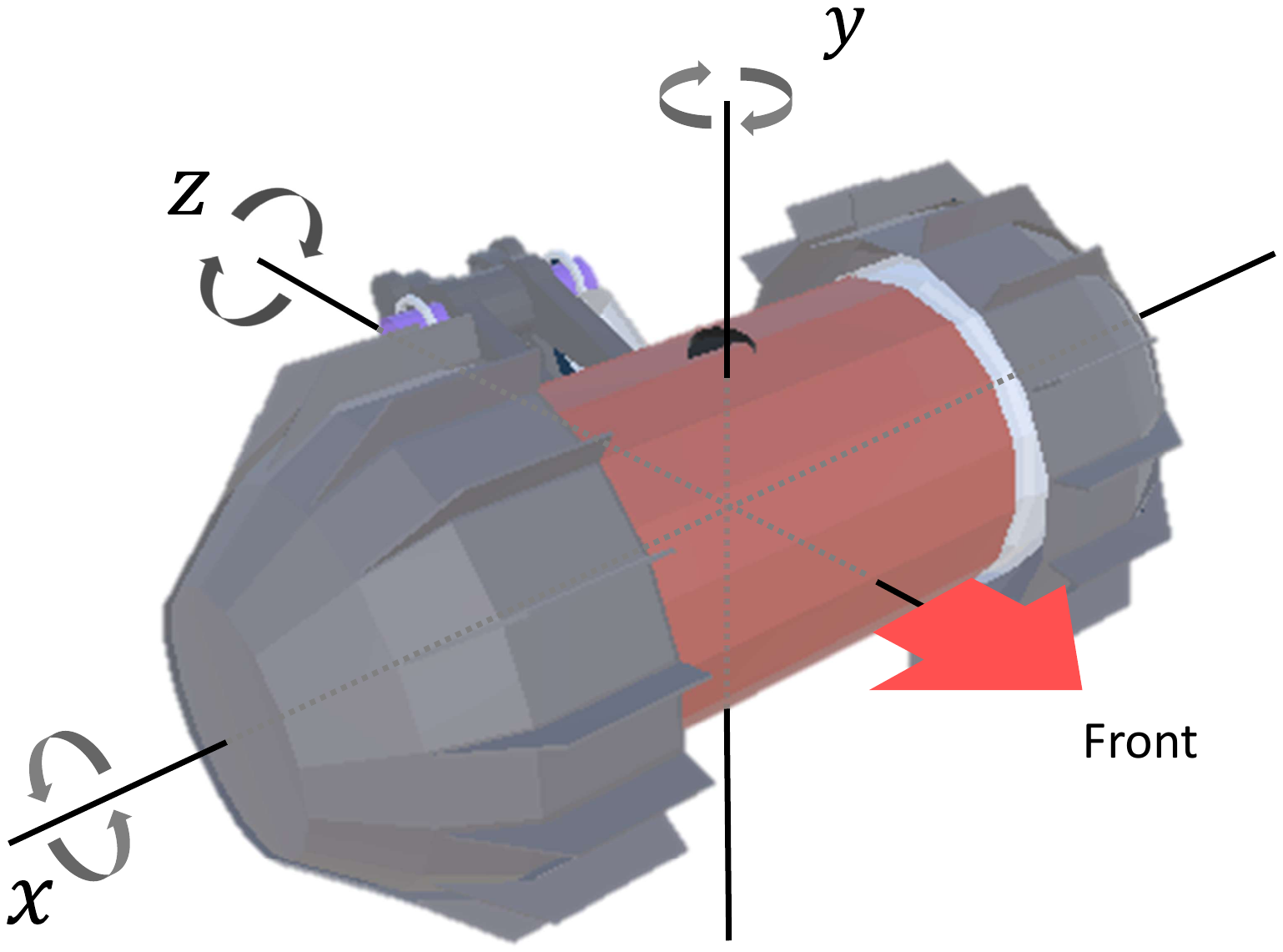}
    \caption{Robot with defined rotation axes}
    \label{fig:posture}
\end{figure}

For the exploration of lava tubes on the lunar surface, two-wheeled differential robots are a promising option because they are cost-effective and easy to transport. Therefore, we used the robot shown in Fig.~\ref{fig:posture}, which can independently actuate its left and right wheels to move forward and backward. It can also observe the environment using equipped sensors. However, in this study, observation data such as the robot's pose (position and orientation) are obtained directly from transforms stored in Unity.

\subsection{Training}
\mypara{Task}
The two-wheeled differential robot is tasked with reaching a target point, which is relatively close to the robot's current position.

\mypara{Training process}
An initial model is trained in the flat area to learn fundamental movements such as moving forward, backward, turning right or left. It is then fine-tuned simultaneously in both the flat and rough areas to obtain a general model, which is used to analyze the type of terrains. The learning process is the same in all the phases (i.e., the initial model and the general model).

\begin{figure}[tbp]
    \centering
    \includegraphics[width=0.9\linewidth]{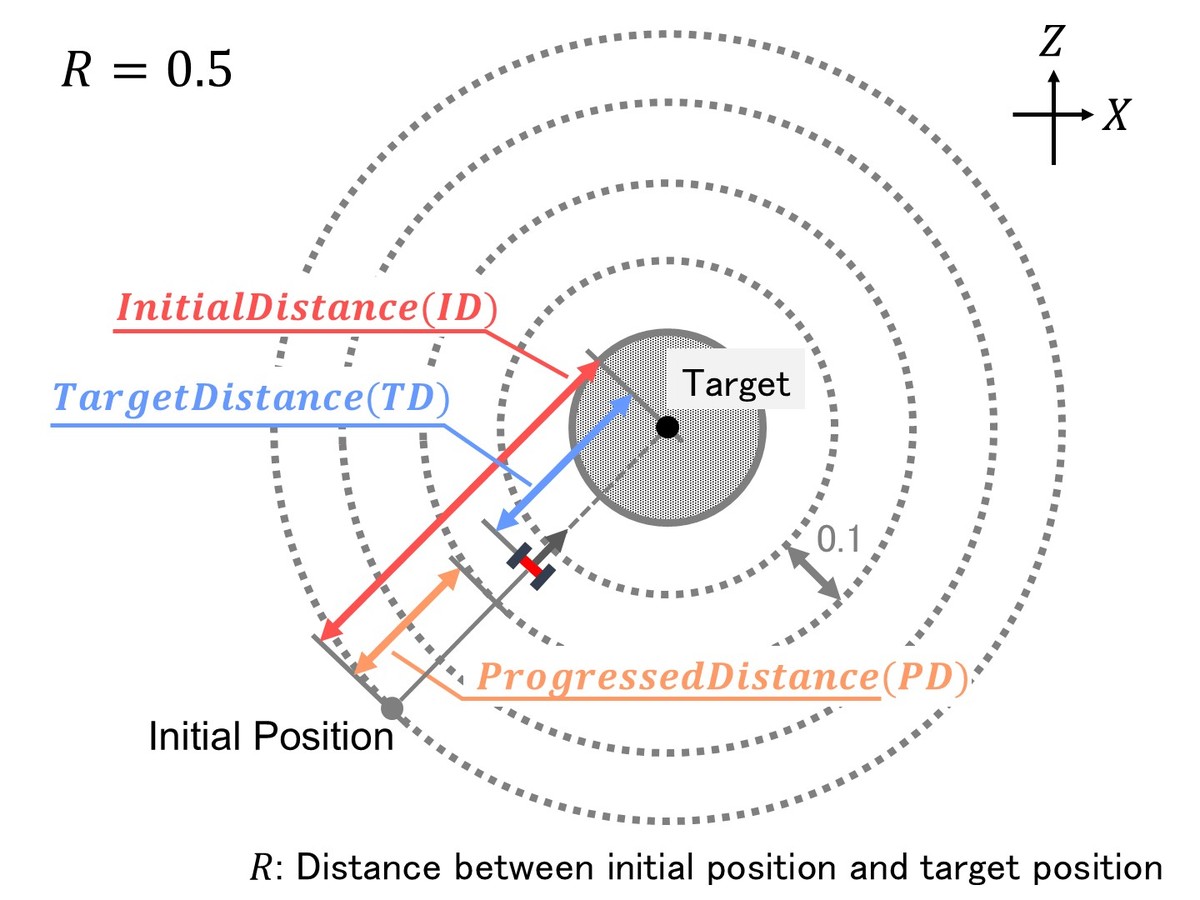}
    \caption{Definitions of the distances}
    \label{fig:def_distances}
\end{figure}

\mypara{Episode}
During training, the robot is first spawned at a random location within the training area. A target is then placed within a circle of radius $5\Delta$ centered at the robot's initial position, where $\Delta$ $= 0.1$. However, to prevent the initial position and the target from being too close, the region within a radius of $\Delta$ centered at the initial position is excluded from the previously defined circle. When the robot reaches the target, the task is considered complete, and the episode terminates. A new episode then begins after a new target is generated within the area defined in Sec.~\ref{sec:environment} centered on the robot's current position, while the robot's pose is not reset between episodes.
If the robot fails to reach its target within the predefined maximum number of steps, $MaxEpisodeSteps (MES)$ (Eq.~\eqref{eq:mes}), the episode ends, and the task is considered failure.
In addition, if the distance between the robot and the target exceeds $PenaltyDistance (PED)$ (Eq.~\eqref{eq:ped}), which changes dynamically during the episode, the episode also terminates. Note that when the distance between the robot and the target becomes less than $\Delta$, shown by the gray-shaded circle in Fig.~\ref{fig:def_distances}, the robot is regarded as having reached the target. 
Here, $MES$ and $PED$ are defined as follows.
\begin{equation}
MES = 300 + 200 ID,
\label{eq:mes}
\end{equation}
where $ID\, (\Delta \le ID \le 5\Delta)$ denotes the distance on the X-Z plane between the robot's initial position and the target. 
\begin{equation}
PED = (ID + 2 \Delta) - k \Delta,
\label{eq:ped}
\end{equation}
where $k$ is the number of times $PED$ is updated during an episode.
Initially $k = 0$ and is increased by 1 each time the robot approaches the target by $\Delta$.

\subsubsection{Observation}
The two-wheeled differential robot observes a total of 10 dimensions: the relative coordinates of the target with respect to its own position (3 dimensions), the distance to the target (1 dimension), and its own pose information (6 dimensions). 

\mypara{The relative coordinates of the target}
Taking the robot's orientation into account, the target coordinates $(t_{x}, t_{y}, t_{z})$ are provided with the current robot's position as the origin.

\mypara{The distance to the target}
The robot obtains the distance on the X-Z plane between its current position and the target position as $d = \sqrt{t_{x}^{2} + t_{z}^{2}}$.

\mypara{3D orientation of the robot}
The robot 3D orientation is provided with 
$(\cos{\theta_{x}}, \sin{\theta_{x}}, \cos{\theta_{y}}, \sin{\theta_{y}}, \break
\cos{\theta_{z}}, \sin{\theta_{z}})$ 
as a part of the observation inputs.
$(\theta_{x}, \theta_{y}, \theta_{z})$ represent the Euler angles around the X-, Y- and Z-axes, respectively.
As shown in Fig.~\ref{fig:posture}, these rotation axes are defined as follows: the X-axis passes through the center of the left to right wheels, the Z-axis runs from the front to the rear of the robot, and the Y-axis is defined as the axis perpendicular to both the X- and Z-axis, with its positive direction determined by a left-hand coordinate system.

\subsubsection{Action}
The robot generates a 2D vector ($A_{left}, A_{right}$) as its action based on the observations. $A_{left}$ and $A_{right}$ represent the desired rotation for the left and right wheels, respectively. These values range from $-3.0$ to $3.0$, and the robot computes the required torque in Unity to achieve them.

\subsubsection{Reward}
In reinforcement learning, the agent (the two-wheeled differential robot) updates its policy to maximize the total reward. The reward used in this training consists of two types: $FinalReward$, given when the robot completes the task and $ProgressReward$, given during navigation. When the robot reaches the target, it receives the $Final Reward$, which is the total value of $BaseReward (BR)$, $OrientationReward (OR)$ and $TimePenalty$ defined as
\begin{equation}
FinalReward = BR + OR - TP.  
\label{eq:finalreward}
\end{equation}
The $BaseReward (BR)$ is defined as $BR = 100 + 10RD$, where $RD = 15 - \sum_{k=1}^{N} k$ and $N = \lfloor ID / \Delta \rfloor$, in order to maintain a balance with $ProgressReward$ described below. Here, the maximum value of $N$ is 5 and thus $RD \geq 0$.
The $OrientationReward (OR)$ is computed as $OR = 50MO$, which promotes the robot maintaining an appropriate posture during navigation, where $MO$ is the episode average of $\cos{\theta_x}$.
Finally, the $TimePenalty (TP)$ is defined as $TP = {steps} / {MaxEpisodeSteps}$, which motivates faster completion of the task, where \textit{steps} is the cumulative number of steps taken until the robot reaches the target. 
Each value plays an important role in learning. The $BaseReward (BR)$ encourages the robot to move toward its target. 

The $ProgressReward$ defined in Eq.~\eqref{eq:progressreward}, on the other hand, uses the $PenaltyDistance (PD)$, which is the Euclidean distance on the X-Z plane between the robot's initial position and its current position, truncated to the first decimal place.
\begin{equation}
Progressed Reward = 100PD,  
\label{eq:progressreward}
\end{equation}
The $ProgressReward$ is given and updated each time the robot approaches the target by $\Delta$. Thus, the moment the robot crosses one of the dashed circles in Fig.~\ref{fig:def_distances} corresponds to the timing at which this reward is received. Each reward point can be obtained only once. For example, if the robot receives the $ProgressReward$ at a distance of 0.3 ($= 3\Delta$), then even if it later moves outside the radius 0.3 circle and returns to 0.3 distance again, it will not receive the reward a second time.

\subsubsection{Hyperparameters}
The hyperparameters used in each stage are shown in Table~\ref{tab:hyperparameters}.
The \textit{Total Steps} represent the number of environment interaction steps used for training. 
The \textit{Batch Size} indicates how many samples are collected before each policy update, and the \textit{Epochs} specify how many times the sampled batch is passed through during a PPO update.
The \textit{Learning Rate} (LR) controls the optimizer's step size, while $\epsilon$ represents the clipping range used in PPO's surrogate objective.
The \textit{Discount Factor} $\gamma$ determines how future rewards are weighted.
The \textit{Entropy} coefficient regulates the strength of entropy regularization to encourage exploration.
GAE~$\lambda$ controls the bias--variance trade-off in Generalized Advantage Estimation.
\textit{Advantage Normalization} indicates whether advantages are normalized before updates.
The \textit{Learning Rate Schedule} determines how the learning rate changes during training.
\textit{Parallel Environments} specify the number of environments executed simultaneously to collect experience.
Finally, \textit{Pretrain} indicates whether the training begins from scratch or from an existing model, where `No' means starting from scratch and `Yes' means starting from an existing model.

\begin{table}[tbp]
\caption{Hyperparameters}
\label{tab:hyperparameters}
\centering
\begin{tabular}{l|rr}
\hline
Hyperparameter & Initial Flat & General \\
\hline
Total Steps & 2.5M & 2.5M \\
Batch Size & 64 & 64 \\
Epochs & 10 & 10 \\
LR & $3\!\times\!10^{-4}$ & $3\!\times\!10^{-4}$ \\
$\epsilon$ & 0.2 & 0.2 \\
$\gamma$ & 0.99 & 0.99 \\
Entropy & $5\!\times\!10^{-4}$ & $5\!\times\!10^{-4}$ \\
GAE $\lambda$ & 0.95 & 0.95 \\
Adv. Norm & Yes & Yes \\
LR Schedule & Linear & Linear \\
Parallel Env. & 9 & 8 \\
Pretrain & No & Yes \\
\hline
\end{tabular}
\end{table}

\vspace{-1mm}
\section{Experiments}
\vspace{-1mm}

\subsection{Observations}
This study aims to determine which terrain features are effective for adaptive policy.
Since the robot is expected to detect its posture using an IMU sensor, we analyzed the corresponding data collected from both flat and rough areas. 
However, although an IMU sensor normally provides noisy inertial measurements that require filtering to estimate orientation, we assumed in this study that the robot can directly access noise-free orientation data, as provided by the transforms in Unity. For capturing the robot's body inclination, we focus on using roll ($\theta_z$) and pitch ($\theta_x$).

\subsection{Data}
\begin{figure}[tbp]
    \centering
    \includegraphics[width=1.0\linewidth]{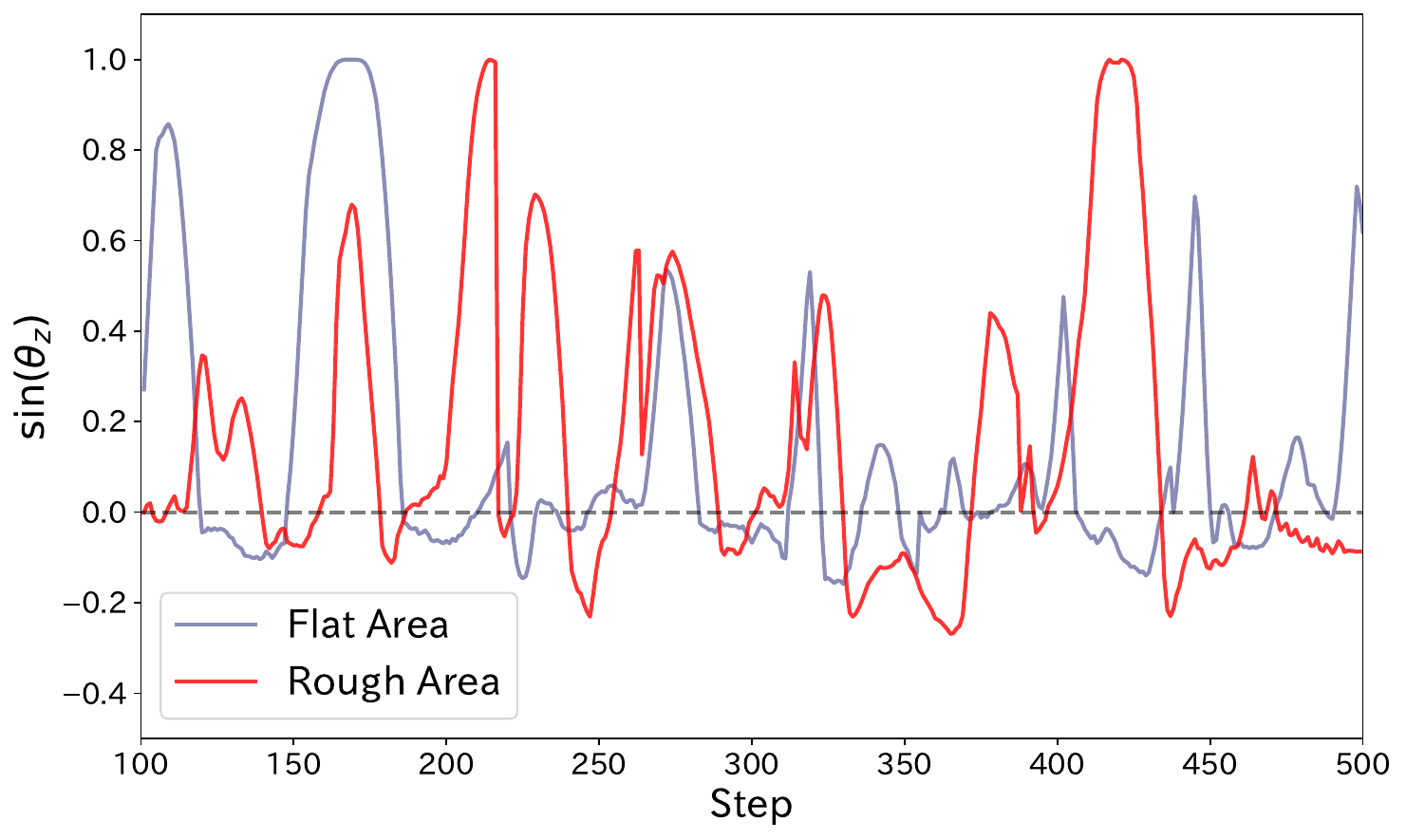}
    \caption{$\sin \theta_z$ (roll) in the two areas}
    \label{fig:sin_roll}
    % \vspace{3mm}
\end{figure}

\begin{figure}[tbp]
    \includegraphics[width=1.0\linewidth]{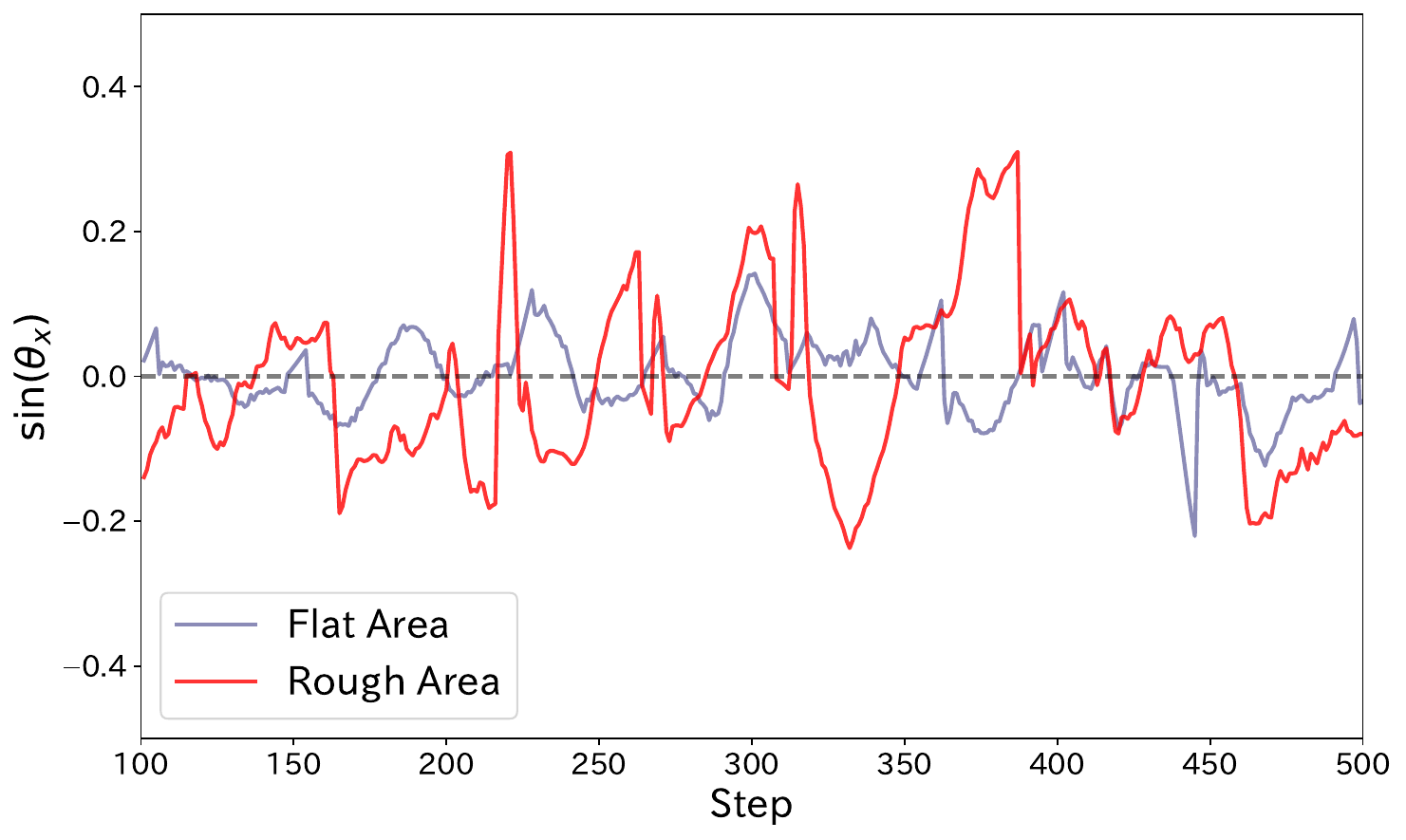}
    \caption{$\sin \theta_x$ (pitch) in the two areas}
    \label{fig:sin_pitch}
    % \vspace{3mm}
\end{figure}

In this experiment, the robot traversed both flat and rough areas defined in Sec.~\ref{sec:environment} and collected its orientation data for 500 steps (0.1~s/step). Since the robot starts the first episode slightly above the ground to implementation constraints, the first 100 steps were discarded.
While collecting these data, the robot performs the task repeatedly.
Figs.~\ref{fig:sin_roll} and \ref{fig:sin_pitch} show the time series of $\sin \theta_z$ and $\sin \theta_x$, respectively, collected in both flat and rough areas.

As can be seen, the difference of the variation in $\sin \theta_x$ (pitch) in the two terrain types appears to be greater than that of $\sin \theta_z$ (roll). This suggests that $\sin \theta_x$ is more informative for capturing terrain features. Therefore, we focus on the $\sin \theta_x$ (pitch) to enable the robot to identify the terrain types it is currently traversing. 

\subsection{Data processing}

\begin{figure}[tbp]
    \centering
    \includegraphics[width=1.0\linewidth]{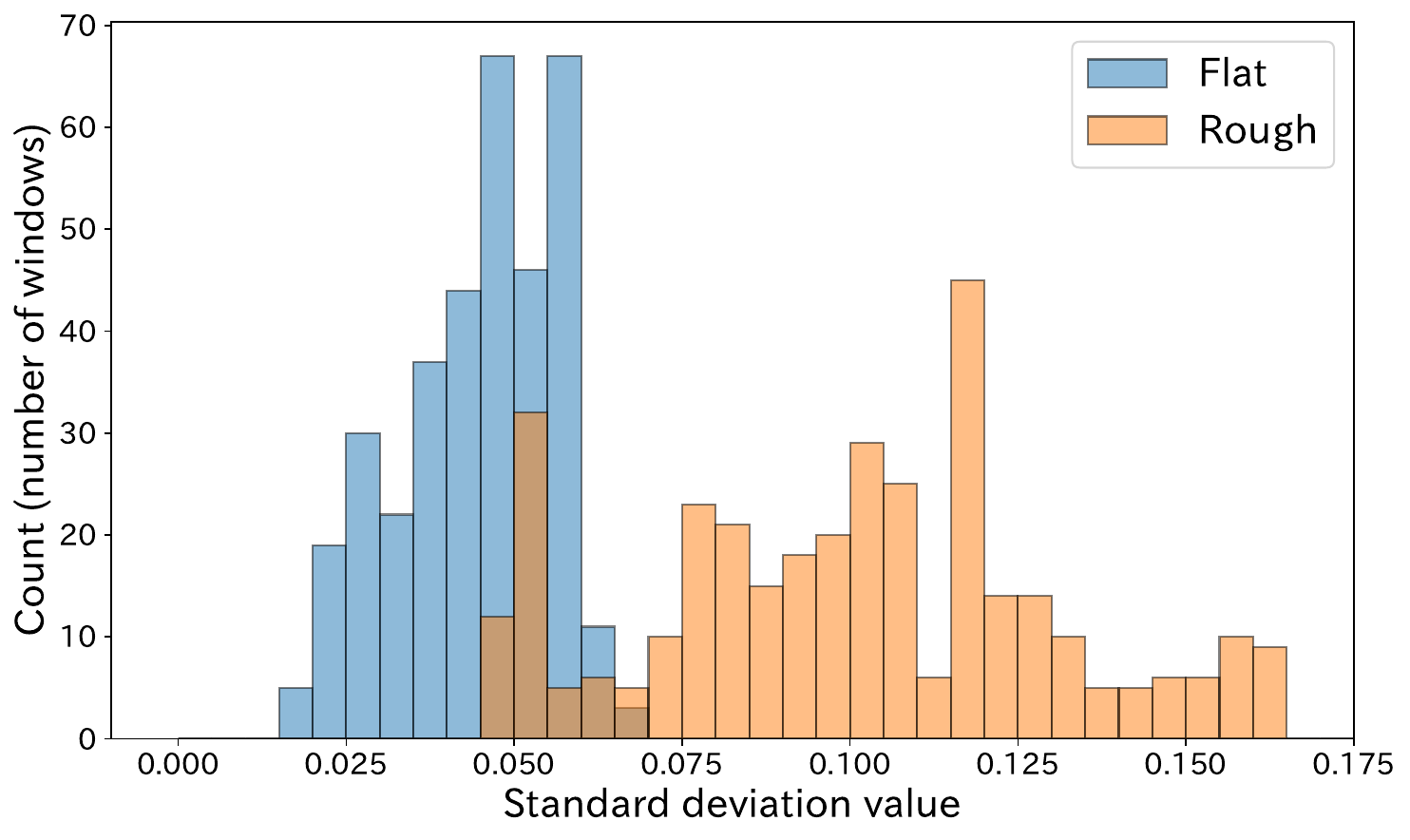}
    \caption{Distribution of $\sin \theta_{x}$ std. (100-step window)}
    \label{fig:dist_sin_pitch}
\end{figure}

To quantify the variability, we use the standard deviation (std.) of the $\sin \theta_x$ values using a rolling window with window size 100. As can be seen in Fig.~\ref{fig:dist_sin_pitch}, the distribution of the $\sin \theta_x$ std. collected in each of the flat and rough areas shows different characteristics. In particular, the distribution of the rough area is clearly shifted to the right compared to that of the flat area and shows a larger spread. 
This indicates that the robot can potentially estimate the terrain type it is currently traversing by analyzing only the recent 100 steps of the $\sin \theta_x$ values.

\subsection{Classifier}

While we collected the $\sin \theta_x$ data separately in the flat and rough areas, in practice, the robot does not have prior knowledge about the terrain type it is currently traversing. Therefore, we need to classify the terrain type based on the collected data without using any labels. To estimate the type of terrain in an unsupervised manner, we adopted Gaussian mixture model (GMM). 

GMM is an unsupervised clustering algorithm that groups given data points without human supervision. Specifically, it assumes that the data are generated from a mixture of several Gaussian distributions with unknown parameters, which can be estimated using the Expectation-Maximization (EM) algorithm.
One advantage of this algorithm is that it can account for the unequal variance observed between groups. As can be seen in Fig.~\ref{fig:dist_sin_pitch}, the std. of each group differs. This requires us to take this difference into account. Unlike K-Means, which assumes equal variance, GMM explicitly models the distinct spread of the flat and rough distributions. 

We combined the two sets of $\sin \theta_x$ data from the flat and rough areas and applied GMM to them. We conducted this experiment while varying the window size, using 10, 20, 40, and 70 steps.
After estimating the parameters of the two Gaussian distributions, we consider them as representing the flat and rough terrain classes and classified each data point.

\subsection{Results}
\begin{figure}[tbp]
    \centering
    \begin{subfigure}[t]{0.49\linewidth}
        \centering
        \includegraphics[width=\linewidth]{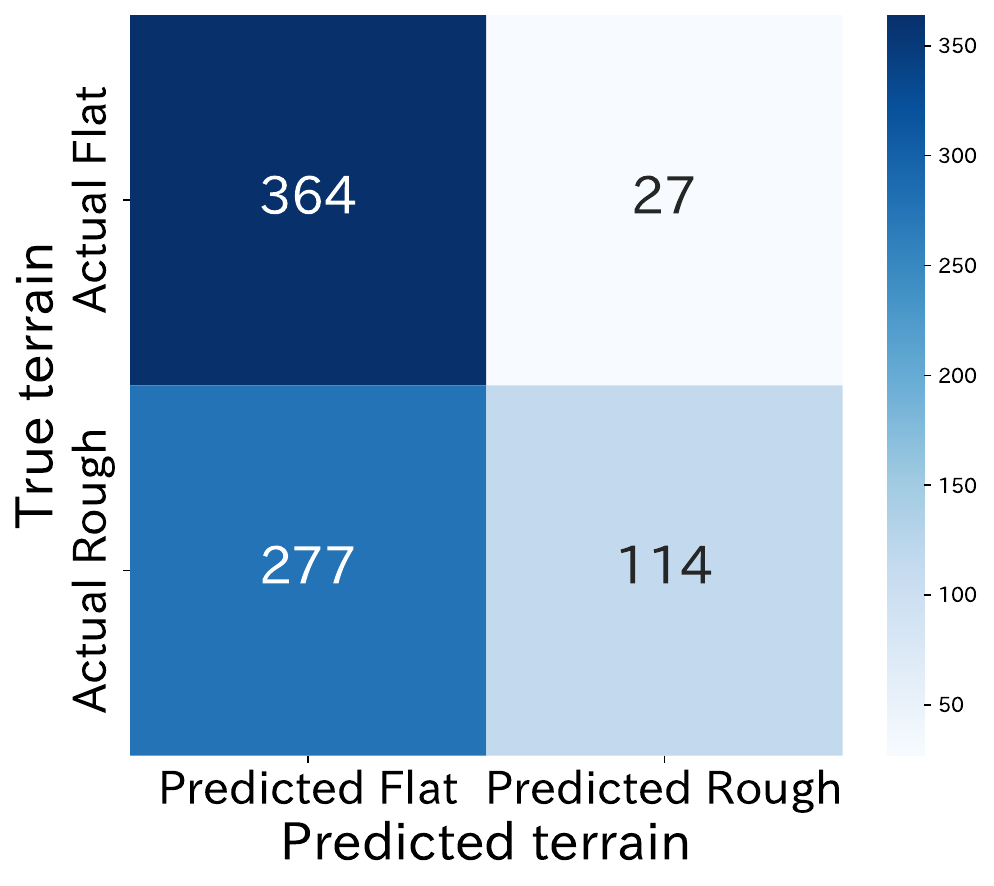}
        \caption{Window size = 10}
        \label{fig:heatmap10}
    \end{subfigure}
    \hfill
    \begin{subfigure}[t]{0.49\linewidth}
        \centering
        \includegraphics[width=\linewidth]{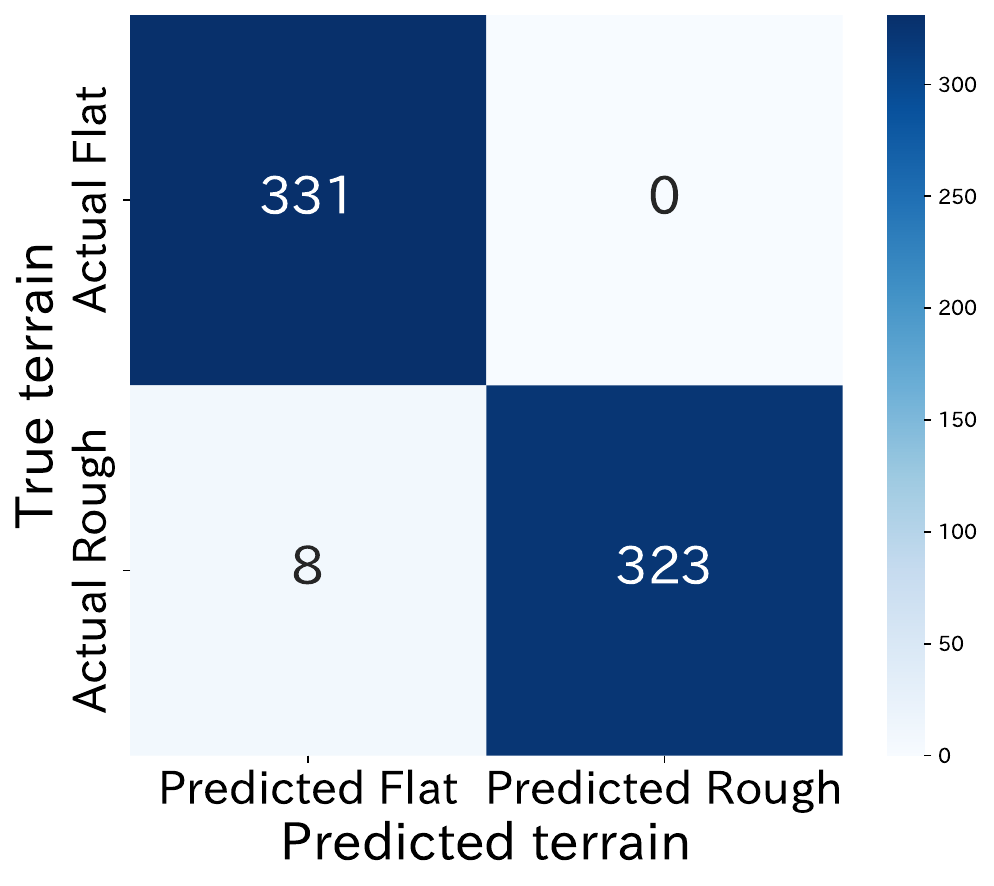}
        \caption{Window size = 70}
        \label{fig:heatmap70}
    \end{subfigure}
    \caption{Cross evaluation heatmaps}
    \label{fig:heatmaps}
    \vspace{-2mm}
\end{figure}

Table~\ref{tab:results} shows the mean and variance for each class, as well as the accuracy of each window size.
The accuracy is computed by comparing the predicted terrain type with the ground-truth labels, i.e., the original terrain labels before combining the data.
The accuracy improves as the window size increases. If the robot considers the most recent 70 steps, it can identify terrain with more than 98\% accuracy.
Figure~\ref{fig:heatmaps} shows the confusion matrices for window sizes 10 and 70.
When the window size is small (10 steps), the classifier frequently misidentifies rough terrain as flat, indicating insufficient information for reliable discrimination. As the window size increases, the separation between flat and rough terrain becomes more distinct.
With a window size of 70 steps, the misclassification rate is significantly reduced, and the classifier correctly identifies most rough-terrain segments. 
These results demonstrate that a larger temporal window stabilizes the std. of the $\sin \theta_x$ and improves the reliability of terrain estimation.

\begin{table}[tbp]
\caption{GMM clustering performance by window size. The middle column displays the estimated mean and standard deviation for the flat / rough clusters, respectively.}
\label{tab:results}
\centering
\begin{tabular}{|c|c|c|}
\hline
Window size & Mean (Std.) & Accuracy \\\hline
10 & 0.02 (0.01) / 0.08 (0.04) & 61.13 \%  \\ \hline
20 & 0.03 (0.01) / 0.08 (0.03) & 70.73 \%  \\ \hline
40 & 0.05 (0.01) / 0.11 (0.07) & 85.87 \%  \\ \hline
70 & 0.05 (0.01) / 0.11 (0.03) & 98.79 \%  \\ \hline
\end{tabular}
\end{table}

\subsection{Discussion}

In this paper, we trained a general terrain model using reinforcement learning, which serves as the basis for subsequent fine-tuning into terrain-specialized models, and evaluated how accurately the robot can estimate terrain types during traversal using this general model. If the robot can capture terrain features with accuracy comparable to that observed in our experiment, it should be able to reliably determine the terrain type and decide which model to train. However, the robot cannot obtain such clean data in real-world settings. Therefore, experiments using an actual IMU sensor are required, and it is necessary to clarify how raw sensor data should be processed for use in adaptive policy switching. In addition, the real lunar environment contains a wider variety of terrain types, meaning that future studies must consider increasing the number of terrain classes.

\section{Conclusion}

We investigated whether a robot can estimate terrain types during traversal using the standard deviation of $\sin \theta_x$ (pitch data). Our evaluation demonstrated that this measure enables reliable discrimination between flat and rough terrain when an appropriate window size is employed, indicating the feasibility of terrain-aware policy switching.

However, the approach relies on clean data provided by the simulator. In real environments, the robot must process noisy IMU measurements and handle a wider variety of terrain types than those considered here. Future work will therefore focus on validating this method with real IMU sensors and extending the classifier to accommodate more diverse terrain conditions expected on the lunar surface. Furthermore, the proposed method should be incorporated into the adaptive policy switching framework and evaluated on an actual robot to assess its effectiveness in navigating diverse terrain conditions.

\section*{Acknowledgment}

This work was supported by JST Moonshot R\&D Program Grant Number JPMJMS2238, Japan.

\end{document}